%% file: root.tex
\documentclass[letterpaper, 10 pt, conference]{ieeeconf}  

\IEEEoverridecommandlockouts                              

\usepackage{graphics} 
\usepackage{epsfig} 
\usepackage{mathptmx} 
\usepackage{times} 
\usepackage{amsmath} 
\usepackage{amssymb}  
\usepackage{bm}
\usepackage{multirow}
\usepackage{cite}

\usepackage{color}

\title{\LARGE \bf
    Bayesian 
    Active Learning for Sim-to-Real Robotic Perception	
}

\author{Jianxiang~Feng$^{1,2}$, Jongseok~Lee$^{1}$, Maximilian~Durner$^{1,2}$ and Rudolph~Triebel$^{1,2}$
\thanks{$^{1}$ Institute of Robotics and Mechatronics, German Aerospace Center (DLR), Wessling, Germany. {\tt\small email: jianxiang.feng@dlr.de}}%
\thanks{$^{2}$ Technical University of Munich (TUM), 80333 Munich, Germany.}
}

\makeatletter
\def\endthebibliography{%
	\def\@noitemerr{\@latex@warning{Empty `thebibliography' environment}}%
	\endlist
}

\begin{document}

\maketitle

\begin{abstract}

While learning from synthetic training data has recently gained an
increased attention, in real-world robotic applications, there are
still performance deficiencies due to the so-called Sim-to-Real
gap. In practice, this gap is hard to resolve with only synthetic data.
Therefore, we focus on an efficient acquisition of real data
within a Sim-to-Real learning pipeline. 
%
%
Concretely, we employ deep Bayesian active learning to minimize manual
annotation efforts and devise an autonomous learning paradigm to
select the data that is considered useful for the human expert to
annotate.  To achieve this, a Bayesian Neural Network (BNN) object
detector providing reliable uncertainty estimates is adapted to infer
the informativeness of the unlabeled data.  Furthermore, to cope with
misalignments of the label distribution in uncertainty-based sampling,
we develop an effective randomized sampling strategy that performs
favorably compared to other complex alternatives.  In our experiments on
object classification and detection, we show benefits of
our approach and provide evidence that labeling efforts can be reduced
significantly.  Finally, we demonstrate the practical
effectiveness of this idea in a grasping task on an assistive robot.

\end{abstract}

\input{chapters/introduction}
\input{chapters/related_work}

\input{chapters/method}
\input{chapters/experiments}

\input{chapters/conclusion}
\input{chapters/acknowledge}






\bibliographystyle{IEEEtran}
\bibliography{IEEEabrv,references}

\end{document}

%% file: chapters/introduction.tex
\section{Introduction}
Over the last years, the performance of computer vision increased sharply, leading to the urge of employing such approaches on robotic vision tasks such as object classification, detection~\cite{lin2017focal, durner_unknown_2021} and pose estimation~\cite{sundermeyer2020augmented}. 
In this context, the necessity of large amounts of annotated, task-related training data is a main issue, particularly for tasks relying on semantic features such as object classification or detection.
Therefore, a compelling solution is to learn from synthetic data. 
Like this, large amount of annotated data can be obtained from simulation with relatively less time and manual efforts~\cite{bousmalis2018using, georgakis2017synthesizing, 10.1109/iros.2017.8202133}. 
With the emergence of open-source image synthesizing pipelines~\cite{denninger2019blenderproc, muller_photorealistic_2021}, this solution becomes even more accessible in practice. 
However, although these pipelines continue improving in fidelity and become more photo-realistic, there are subtle but important differences between simulation and real domain.
This leads to the so-called \textit{Sim-to-Real gap} which is the main barrier to transfer this technique to real world robotic perception.
Several works address this gap by applying techniques such as Domain Randomization (DR)~\cite{10.1109/iros.2017.8202133, sundermeyer2020augmented} and Domain Adaptation (DA)~\cite{bousmalis2018using, Tanwani_DIRL_CORL_20} with certain improvements.
Yet, the unpredictable variability of real-world scenes prevents a complete elimination of the reality gap~\cite{electronics10121491}.

\begin{figure}[t!]
	\centering
	\includegraphics[height=5.8cm,width=\linewidth]{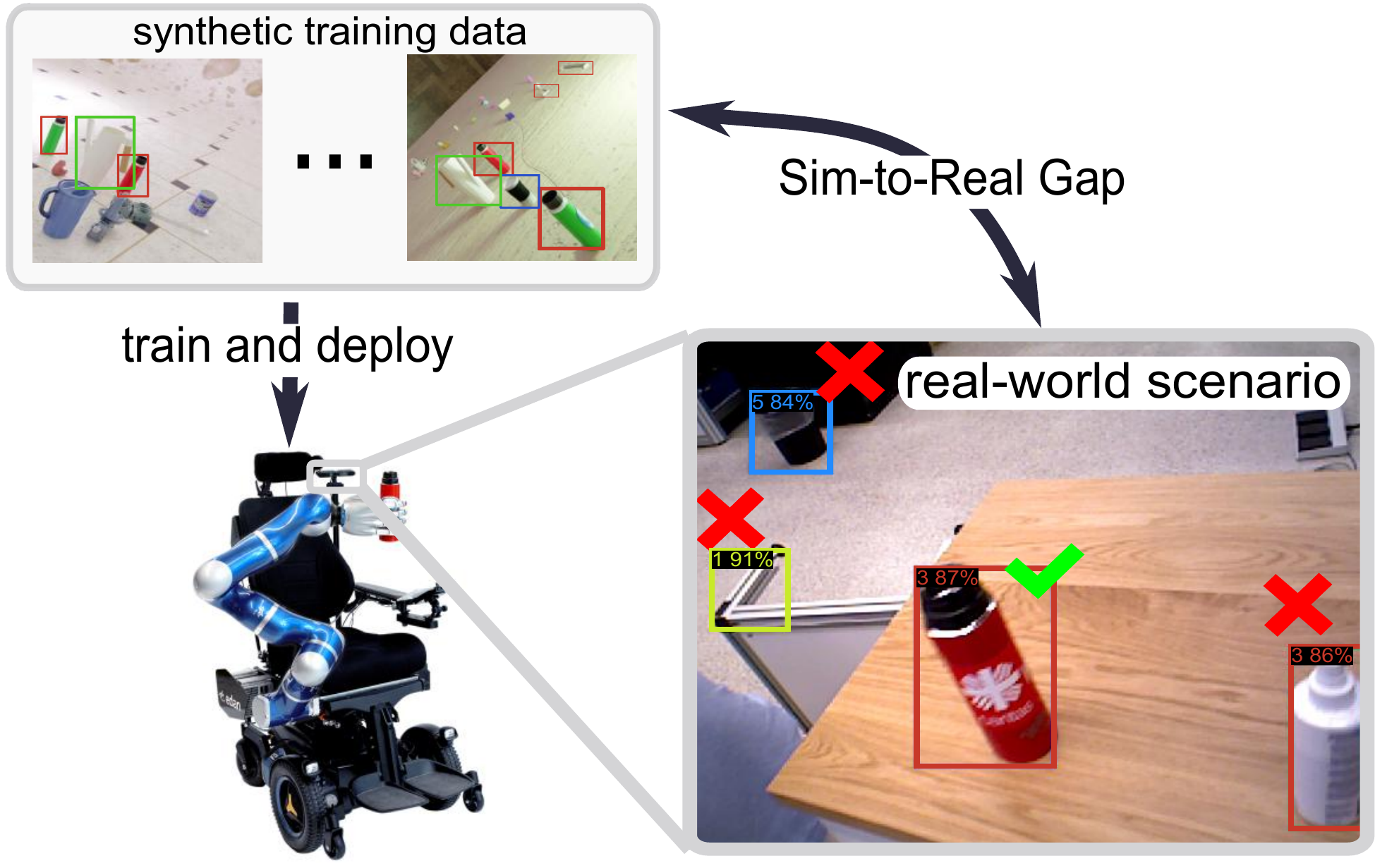}
	\caption{\textbf{Illustration of a practical problem.} Deploying a detector trained with only synthetic images on real-world scenarios leads to under-performances. These inaccuracies (denoted by red crosses) such as false positives are due to the Sim-to-Real gap and for this, a few informative real images can improve the performance. Therefore, this work investigates the question on how to collect such informative real images via active learning.} 
	\label{fig:teaser1}
	\vspace*{-5mm}
\end{figure}

We encounter similar issues in our real lab environment, when deploying an object detector~\cite{lin2017focal} trained on photo-realistic images on our robotic platform EDAN~\cite{vogel2020edan}.
From our practical experience, variables such as sensor characteristics, illumination, or textures cannot be modeled to precisely match the real environments.
Even after a careful tuning of DR, we find that the object detector fails to generalize well in the real-world scenario (e.g., clutter in lab environment see Fig. \ref{fig:teaser1}).
To overcome this, we applied domain-oriented fine-tuning, by using real data of the underlying robotic application, like~\cite{durner_experience-based_2017}.
Hence, based on our use-case, we advocate that real data is indispensable for a robot to robustly adapt from simulation to real world.

This however, comes with the requirement of tedious, time consuming manual labeling.
In this work, we investigate on the question: \textit{How to bridge the Sim-to-Real gap with minimum annotation efforts?}
Having a model trained on synthetic images, we propose an Active Learning (AL) pipeline that can efficiently bridge the still present Sim-to-Real gap.
In contrast to our previous work~\cite{feng2019introspective}, we here aim for autonomous acquisition of as few annotated real images as possible.
Based on a deep Bayesian Active Learning (AL) framework\cite{gal2016dropout,harakeh2020bayesod}, we analyze different strategies to select the most informative data samples.
Further we devise a simple yet effective strategy to mitigate the lack of diversity in the selected data, caused by the label distribution shift between simulation and real domain~\cite{prabhu2019sampling, zhao2021active}.
Note, that the latter is important for performance gain in AL under domain shift (simulation vs. reality domain in our case)\cite{su2020active, prabhu2021active}.
Moreover, for the more challenging 2D object detection task, we suggest to incorporate regression uncertainty into the selection process due to its multi-task characteristic (including both classification ($cls$) and regression ($reg$)).  

Concretely, we train a BNN with synthetic images, which can be obtained from another task-relevant data set or generated by photo-realistic image synthesizers. 
In a second step, a pool of task-specific real images are forwarded to the model.
According to the scores from the acquisition function and a sampling strategy, a small subset of samples is selected and solicited for human annotations.
The labeled data is then used to adapt the model.
The aforementioned procedures can be repeated iteratively until the desired performance is achieved.

Besides the empirical validation of the proposed idea on a classification task,
we then conduct evaluation on two more challenging 2D object detection data sets, one with large Sim-to-Real domain shift and another with less to show that the proposed idea can help bridge the gap in a cost-effective way, significantly better than the random baseline and competitive against the state-of-the-art approaches. 
In addition, we provide a failure case on a third object detection data set to help identify the working scenarios of the proposed idea.
To demonstrate the practicality and effectiveness, we further deploy the pipeline on our real robot and show a significantly positive impact of the visual perception within grasping as downstream task.

In summary, the main contributions of this work are:
\begin{itemize}
	\item we propose to actively and efficiently close the Sim-to-Real gap by applying a BNN in an Active Learning (AL) framework.
	\item we introduce a simple yet effective sampling strategy to mitigate the label distribution shift in Bayesian AL under domain shift.
	\item we conduct experiments to empirically show the positive impact of the proposed pipeline on both object classification and 2D object detection tasks, clearly outperforming the random baseline and closely competing against the state-of-the-art approaches 
	\item we demonstrate the applicability in reducing labeling efforts on a real robotic system.
\end{itemize}

Importantly, the accompanying video provides qualitative results including the demonstration on an assistive robot.
The code of the implementation will be publicly available\footnote{https://github.com/DLR-RM}.

%% file: chapters/related_work.tex
\newcommand{\etal}{\textit{et al.}~}
\section{Related Work}

\paragraph{Sim-to-Real Transfer}
Sim-to-Real transfer is mainly tackled with DR and DA. 
The former treats the real test scenario as one instance of many synthetic ones generated by randomizing the parameters of the synthesizer such as materials, lightening, backgrounds, and plausible geometric configurations ~\cite{hinterstoisser2018pre, Photorealistic_Image_Synthesis}.
In contrast, DA focuses on learning domain-invariant representations across the different domains (e.g. synthetic and real domain in this context) by sometimes including data of the target domain~\cite{bousmalis2018using}.
Though DA has achieved impressive performance, as mentioned by different researchers, when only relying on unlabeled data, the domain gap is hard to diminish both in theory \cite{Tanwani_DIRL_CORL_20} and in practice \cite{zhu2019adapting, chen2018domain}.
Considering this issue, the paradigm of active learning is appealing to address the reality gap by utilizing annotated real data in an efficient way.
In pool-set based active learning~\cite{cohn1996active}, the aim is to reach certain level of performance with as less data as possible. 
In case of supervised learning, the data is selected based on their informativeness, which can be measured by different quantities such as the output uncertainty, the disagreement of a committee, or the expected model change \cite{feng2019deep, kirsch2019batchbald}.
We also stress that active learning is complementary to the aforementioned techniques.
While recent works such as ~\cite{su2020active, prabhu2021active} argue for the fusion of DA and active learning to obtain better performance, we additionally use DR in this work.
Nevertheless, none of them considers employing BNNs for this purpose and most of them focus on classification tasks, which are less relevant for the robots in the real world. 
Wen \etal \cite{wen2019bayesian} apply BNNs for DA, but they only focus on conventional passive learning paradigm and classification tasks.
We aim to study the active learning paradigm for Sim-to-Real transfer on a more challenging real-world object detection task, which is arguably more relevant for various use-cases of the robots. 

\paragraph{Active Learning for Object Detection}
In the context of active learning for object detection, specific metrics related to characteristics of the underlying network can be applied~\cite{aghdam2019active}.
While in~\cite{roy2018deep} the margin of the bounding box scores in different layers is used, Kao \etal \cite{kao2018localization} consider the localization tightness and stability.
Meanwhile, uncertainty based approaches~\cite{feng2019deep, choi2021active, prabhu2021active} are also able to achieve competitive performances in the field of object detection. 
Most of uncertainty based approaches are built on BNNs~\cite{gal2016dropout, lee2020estimating} which can produce more reliable uncertainty estimates.
Along with its theoretic soundness, the task-agnostic characteristic of these approaches can facilitate wider applicability for different fields.
While some only exploit the classification branch for the uncertainty estimation~\cite{miller2018dropout, lee2022}, others~\cite{harakeh2020bayesod} consider both classification and regression branches.
Yet, they rely on larger amount of annotated real world data to initialize the training of the model and update the model in each iteration, while we assume relatively small amount of real data.

%% file: chapters/method.tex
\section{Problem Formulation and Overview}

We consider two domains: the simulation domain $S$ and the real domain $R$. 
In $S$, we assume the availability of annotated data set, i.e., given the synthetic data $\mathbf{x}^{S}$ and annotated labels $\mathbf{y}^{S}$, we denote the synthetic data set as $\mathcal{D}_S = \{(\mathbf{x}_i^{S}, \mathbf{y}_i^{S})\}_{i=1}^{N_S}$ where $N_S$ is the number of data points. 
In contrary, $R$ contains an unlabeled data set $\mathcal{D}_T = \{(\mathbf{x}_i^{R})\}_{i=1}^{N_R}$ which constitutes of $N_R$ number of real images $\mathbf{x}^{R}$. 
We further extend the notations to define an object detection task including classification ($cls$) and regression ($reg$) tasks. 
\begin{figure}[ht]
	\centering
	\includegraphics[width=1\linewidth]{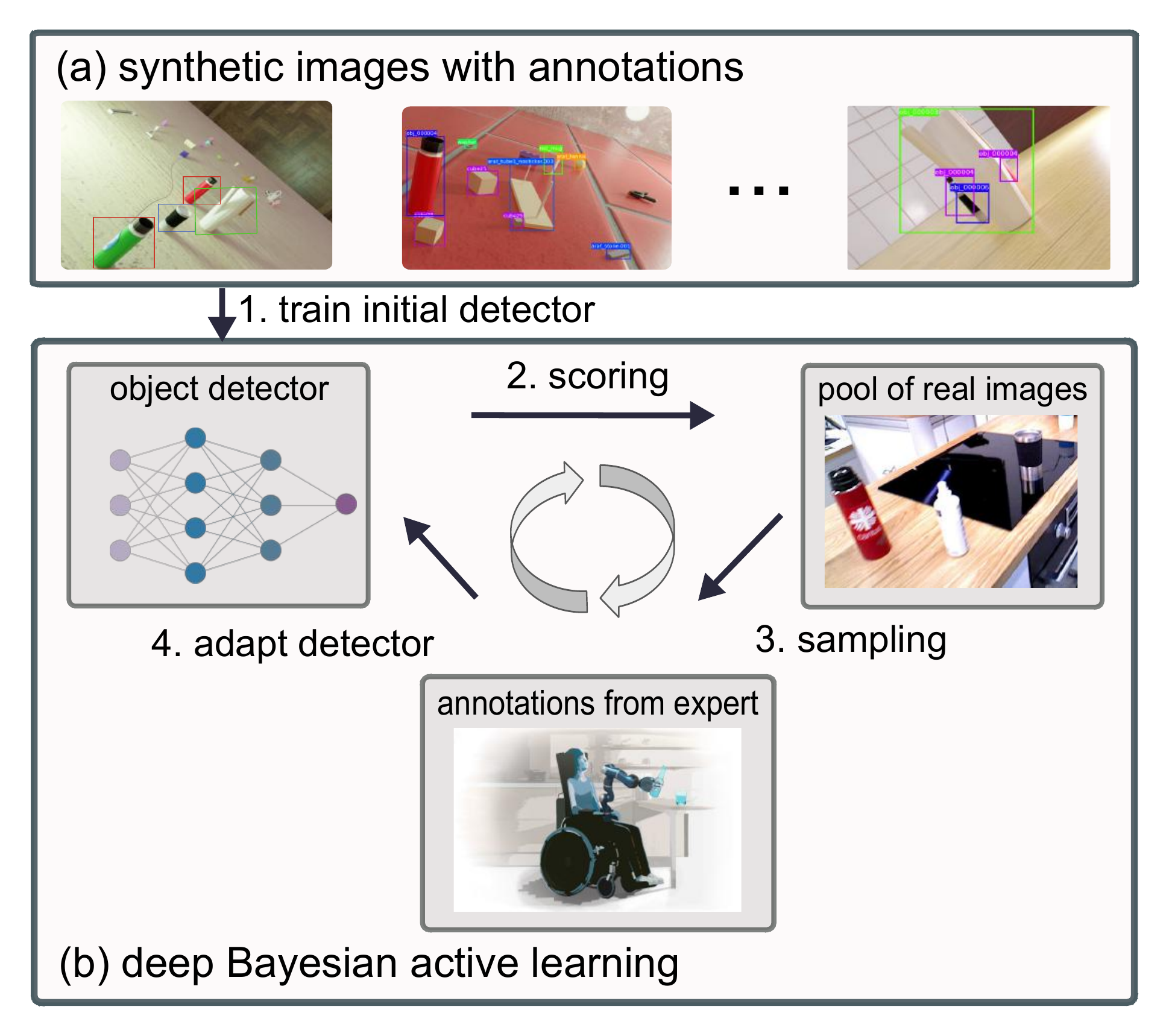}
	\caption{\textbf{The proposed Sim-to-Real pipeline.} Using labeled synthetic images, we first train an initial BNN object detector. Then, we rely on deep Bayesian AL to select the most informative images from a pool of unlabeled real images. The scoring of all the images in the pool is obtained via an acquisition function, while sampling is applied to deal with the foreground class imbalance problem. Based on the selected images, the human expert performs the annotation and the detector is adapted via fine-tuning. The process is repeated to close for Sim-to-Real transfer.} 
	\label{fig:teaser2}
	\vspace*{-5mm}
\end{figure} 
Given the space of inputs $\mathcal{X}$ (both synthetic and real images) and outputs $\mathcal{Y}$ (sets of object classes $\mathbf{c}$ and their 2D location as bounding boxes $\mathbf{b}$), we define the object detector as a function $\mathcal{M}_{\mathbf{\theta}}:\mathcal{X}\to\mathcal{Y}$ with parameters $\mathbf{\theta}$. 
Naturally, our objective is to obtain an object detector in the real domain $R$, for which synthetic data $\mathcal{D}_S$ can be exploited. 

To achieve this goal, the proposed pipeline (depicted in Fig. \ref{fig:teaser2}) relies on deep Bayesian AL. 
What motivates our approach is that in practice, this so-called Sim-to-Real transfer can be achieved by combining (a) the large amounts of annotated synthetic data, and (b) a few but the most informative real images with annotations from human expert. 
Importantly, we conjecture that the real images can bridge the reality gap in a simple and effective manner, and thus, this work focuses on reducing the amounts of needed real images. 
For this, as shown in Fig. \ref{fig:teaser2}, (i) we use $\mathcal{D}_S$ to train an initial model with domain randomization. 
(ii) Then, treating the unlabeled real data $\mathcal{D}_T$ as a pool set $\mathcal{D}_{pool}$, we rank the informativeness of each images with an acquisition function $\mathcal{A}(\cdot)$ and then 
(iii) apply a sampling strategy to create the subset. 
(iv) The labels of this subset is queried to a human expert for manual annotation. 
This process can be repeated for multiple times until the reality gap is diminished. 
Next, we describe and motivate these steps in detail.

\section{The Proposed Pipeline}
This section describes our pipeline of Sim-to-Real transfer for 2D object detection. 
The main components are a BNN object detector for uncertainty quantification (Sec. \ref{sec:4:1}), and deep Bayesian AL framework (Sec. \ref{sec:4:2}).


\subsection{Bayesian Neural Networks for Object Detection}
\label{sec:4:1}

We choose to model the object detector $\mathcal{M}_{\mathbf{\theta}}$ as a BNN, in order to obtain its uncertainty estimates. 
BNNs achieve this by reasoning about the model uncertainty, which indicates \textit{what the model does not know}. 
Reasoning about the model uncertainty, the AL framework can later leverage this information to label the most uncertainty data to the model itself. 
To do so, given the training data $\mathcal{D}_{train}$ and a test data sample $\mathbf{x}^\ast$, BNNs produce the output distribution $p(\mathbf{y}^\ast\mid \mathbf{x}^\ast, \mathcal{D}_{train})$ by marginalizing over the models' distribution:
\begin{equation}
\label{eq:pred_dist}
p(\mathbf{y}^\ast\mid \mathbf{x}^\ast, \mathcal{D}_{train} ) = 
\int p(\mathbf{y}^\ast\mid \mathbf{x}^\ast,\mathbf{\theta}) p(\mathbf{\theta} \mid \mathcal{D}_{train} ) d \mathbf{\theta}.
\end{equation}
In \eqref{eq:pred_dist}, $p(\mathbf{y}^\ast\mid \mathbf{x}^\ast,\mathbf{\theta})$ is the observation likelihood, and $p(\mathbf{\theta} \mid \mathcal{D}_{train}$ is the distribution over the weights $\mathbf{\theta}$. As a closed form solution to the integral in \eqref{eq:pred_dist} does not exist, the Monte-Carlo integration is often used for a numerically solution \cite{gawlikowski2021survey}. As a note, our AL pipeline uses both the synthetic and the annotated real images as the training set $\mathcal{D}_{train}$, and the new images $\mathbf{x}^\ast$ are samples from the pool set $\mathcal{D}_{pool}$.


However, applying BNNs to the existing anchor-based detectors such as Retinanet \cite{lin2017focal} requires several adaptations \cite{miller2018dropout, harakeh2020bayesod}. This is due to their post-processing steps, i.e., (i) \textit{miss-correspondence between the anchor predictions and final outputs}, and (ii) \textit{hard cut-off behavior in non-maximum suppression (NMS) step}. For these, the BayesOD framework \cite{harakeh2020bayesod} can be used, which performs Monte-Carlo sampling for each anchor prediction before NMS steps, and relies on Bayesian inference to infer the output distributions. Intuitively, BayesOD clusters outputs in anchor level using spatial affinity. To explain, assume that such cluster contains $M$ anchors and consider the highest classification score as center of this cluster (indexed by 1). The other outputs are considered as measurements to provide information for the center, denoted by $\hat{\mathbf{c}}_i$ and $\hat{\mathbf{b}}_i$. By further denoting the final predictive distributions for $cls$ and $reg$ of this cluster as $p_{[\hat{\mathbf{c}}_1, ..., \hat{\mathbf{c}}_M]}(\mathbf{c}|\mathbf{x}^\ast, \mathcal{D}_{train})$ and as $p_{[\hat{\mathbf{b}}_1, ..., \hat{\mathbf{b}}_M]}(\mathbf{b}|\mathbf{x}^\ast, \mathcal{D}_{train})$ respectively, the final output distributions are computed as:
\begin{equation*}
\label{eq:bod_cls}
p_{[\hat{\mathbf{c}}_1, ..., \hat{\mathbf{c}}_M]}(\mathbf{c}|\mathbf{x}^\ast, \mathcal{D}_{train}) \propto 
p_{\hat{\mathbf{c}}_1}(\mathbf{c}|\mathbf{x}^\ast, \mathcal{D}_{train}) \prod_{i=2}^{m}p(\hat{\mathbf{c}}_i|\mathbf{c}, \mathbf{x}^\ast, \mathcal{D}_{train}),
\end{equation*}
\begin{equation*}
\label{eq:bod_reg}
p_{[\hat{\mathbf{b}}_1, ..., \hat{\mathbf{b}}_M]}(\mathbf{b}|\mathbf{x}^\ast, \mathcal{D}_{train}) \propto 
p_{\hat{\mathbf{b}}_1}(\mathbf{b}|\mathbf{x}^\ast, \mathcal{D}_{train}) \prod_{i=2}^{m}p(\hat{\mathbf{b}}_i|\mathbf{b}, \mathbf{x}^\ast, \mathcal{D}_{train}).
\end{equation*}
Here, $p_{\hat{\mathbf{c}}_1}(\mathbf{c}|\mathbf{x}^\ast, \mathcal{D}_{train})$ represents the per-anchor predictive distribution of the cluster center, while $\prod_{i=2}^{m}p(\hat{\mathbf{b}}_i|\mathbf{b}, \mathbf{x}^\ast, \mathcal{D}_{train})$ is the likelihood of each cluster member given the output. 
When we choose the Gaussian and Categorical distributions for $cls$ and $reg$ tasks respectively, the sufficient statistics of them such as mean and covariance matrix can be computed analytically. 
We refer  more details in \cite{harakeh2020bayesod} and next, we discuss the AL framework that relies on the BayesOD framework.
\subsection{Bayesian Active Learning for Sim-to-Real}
\label{sec:4:2}
With the uncertainty estimates of an object detector, the AL pipeline needs to choose the images for annotation.
This selection of images is done via an acquisition function. 
Moreover, due to the domain shift between $S$ and $R$, a sampling strategy also needs to be devised to mitigate the bias in the selected data set.
We describe below these components and our design choices.

\subsubsection{Acquisition Function}

We define the acquisition function based on the uncertainty estimates from the BNN detector. 
In this step, the acquisition function is used to obtain the informativeness scores for each detected instance on one image, and then \textit{aggregated} into one final score to represent the informativeness of the entire image. 
Once the scores are obtained for all the images in the pool set $\mathcal{D}_{pool}$, we sample a subset of them for annotation (\ref{sampling}) in order to adapt the model. 
Specifically, we consider uncertainty from both \textit{category classification} and \textit{bounding box regression}, which are referred to as semantic and spatial uncertainty respectively \cite{hall2020probabilistic}. 
For the semantic uncertainty of the $j$-th detection instance on an image, given the Shannon Entropy measure $\mathcal{H}(\cdot)$, the $cls$ acquisition function $\mathcal{U}_{j, cls}$ is modeled with a Bernoulli distribution as:
\begin{equation}
\label{cls_ent}
\begin{split}
\mathcal{U}_{j, cls} & = \sum_{i=1}^{|\mathcal{C}|}\mathcal{H}(p(c_i|\mathbf{x}^{\ast}, \mathcal{D}_{train})), \\
				& = \sum_{i=1}^{|\mathcal{C}|} [-p(c_i|\mathbf{x}^{\ast}, \mathcal{D}_{train})\log{p(c_i|\mathbf{x}^{\ast}, \mathcal{D}_{train})}\\ 
				& -(1-p(c_i|\mathbf{x}^{\ast}, \mathcal{D}_{train}))\log{(1-p(c_i|\mathbf{x}^{\ast}, \mathcal{D}_{train}))}] .
\end{split}
\end{equation}
In \eqref{cls_ent}, the steps follows from the definition of the entropy, and optimizing the given measure is equivalent to maximizing the information gain \cite{mackay1992information} or information content.

The uncertainty from regression is defined as differential entropy of $p(\mathbf{b}|\mathbf{x}^{\ast}, \mathcal{D}_{train})$ which is approximated by a multivariate Gaussian with covariance matrix $\mathbf{C}_{b}$ calculated from the samples of predicted bounding boxes: 
\begin{equation}
\label{reg_ent}
\begin{split}
\mathcal{U}_{j, reg} & = \mathcal{H}(p(\mathbf{b}|\mathbf{x}^{\ast}, \mathcal{D}_{train}))\\
					  & = \frac{k}{2} + \frac{k}{2}\ln{(2\pi)} + \frac{1}{2}\ln{(|\mathbf{C}_{b}|)},
\end{split}
\end{equation}
where $k$ is the dimensionality of random variable $\mathbf{b}$. Again, this regression acquisition function $\mathcal{U}_{j, reg}$ follows from the definition of entropy for Gaussian distributions, and represents the information content of an image.

We choose to exploit these two quantities by a combination function $comb(\cdot)$, in order to produce the uncertainty score for each of $N_k$ detected instance on $k$-th image. Then, the acquisition function for $k$-th image $\mathcal{A}$ is defined by aggregating scores with a function $agg(\cdot)$ denoted by:
\begin{equation}
\label{int_acq}
	\mathcal{A}(\mathbf{x}_k) = agg_{j\in N_k}(comb(\mathcal{U}_{j, cls}, \mathcal{U}_{j, reg})),
\end{equation}
The combination function $comb(\cdot)$ can be a weighted sum ($sum$) or maximum ($max$) operation \cite{choi2021active}.
The aggregation function $agg(\cdot)$ can be a maximum ($max$), summation ($sum$) or average ($avg$) operation \cite{roy2018deep}. 
What motivates this is the problem itself, i.e. object detection involves both $cls$ and $reg$ tasks and multiple instances in one image.

\subsubsection{Sampling Strategy}
\label{sampling}

\begin{figure}[ht]
	\centering
	\includegraphics[height=3.8cm,width=1\linewidth]{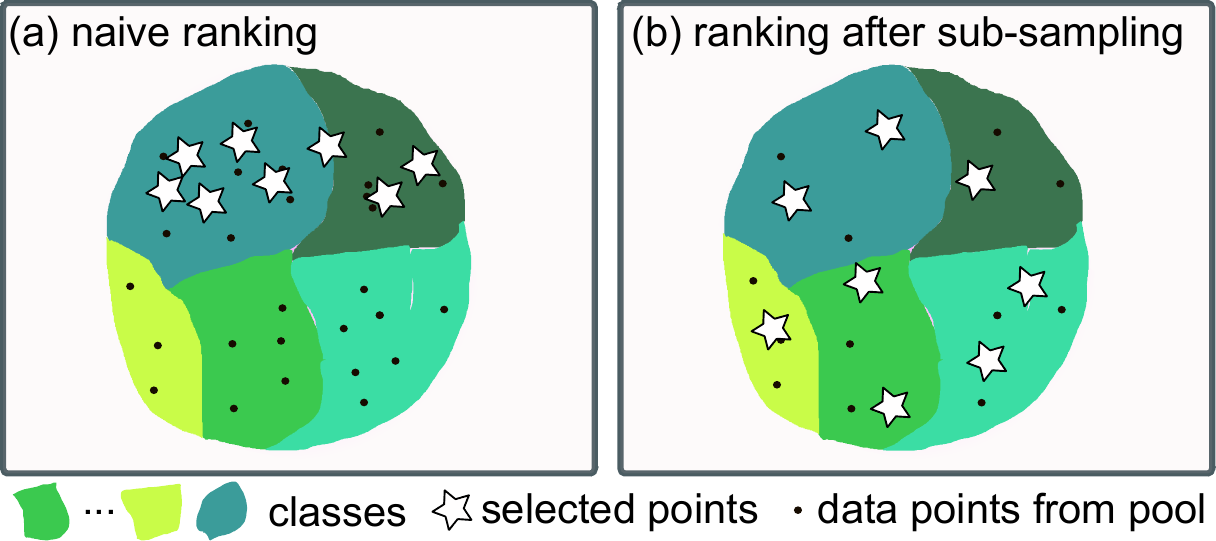}
	\caption{\textbf{Sub-sampling Strategy.} We illustrate the ranking after sub-sampling strategy. A naive ranking selects the most informative points from a few classes of pool data, while the ranking after sub-sampling enables to evenly select the most informative points across the variety of classes. This mitigates the class imbalance problem of AL for object detection, and introduced diversity can improve the performance.}
	\label{fig:subsampling} 
	\vspace*{-4mm}
\end{figure}
One problem in the naive TopN sampling motivates us to combine the TopN sampling with the popular sub-sampling technique \cite{yang2010ensemble}. 
The problem is the violation of the assumption that the simulation domain $S$ and the real one $R$ are the same, which does not hold in fact.
This will further lead to a performance degradation for both AL and object detection training \cite{aggarwal2020active, oksuz2020imbalance}. 
More specific, to select the $B$ most informative images scored based on the model trained on $S$ will result in an imbalance problem in the selected data set. 
Since the algorithm queries only images from real domain $R$, we attribute the under-performance during AL to the label distribution shift \cite{prabhu2019sampling}.
To explain, we denote the distribution followed by sub-sampling as $P_{ss}$($\mathbf{c}$, $\mathbf{b}$) and the distribution 
followed by uncertainty sampling as $P_{unc}$($\mathcal{A}(\mathbf{c}, \mathbf{b})$), which can be a product of delta distribution 
with probability mass placed at the top $B$ scored predictions.
Therefore, the selected data during AL follow the a label distribution $P_{ss}P_{unc}$.
Additionally, we use $P_r$($\mathbf{c}$, $\mathbf{b}$) for the real label distribution, which is assumed to be uniform. 
The goal is to adapt the model with data points drawn from $P_r$, which is unavailable for unlabeled data.
Instead we adapt the model with data points drawn from $P_{ss}P_{unc}$, which ideally should be aligned with $P_r$.
Unlike classification case, in which the label distribution lies in a discrete finite space and importance weighting correction \cite{zhao2021active} can be easily adapted, the label space for object detection is more complex when there is an additional regression task involved.  
The trade-off between alleviation of label distribution shift and utilization of information contained in the uncertainty estimates is thus determined by the distribution form of $P_{ss}$ and the amount of data to be sub-sampled.
Intuitively, by assuming there is certain degree of redundancy in the data set, we select the uniform distribution for $P_{ss}$, which works empirically well, shown in the experiments. 
In practice, the pool set data is filtered by $P_{ss}$ first, and then with $P_{unc}$, the learner thus can choose by considering the informativeness in the sub-sampled data. 
An illustrative explanation on the class imbalance problem, one instance of label distribution shift, is shown in Fig. \ref{fig:subsampling}. 


%% file: chapters/experiments.tex
\section{Experiment}
\begin{figure}
	\begin{center}
		\includegraphics[width=1\linewidth]{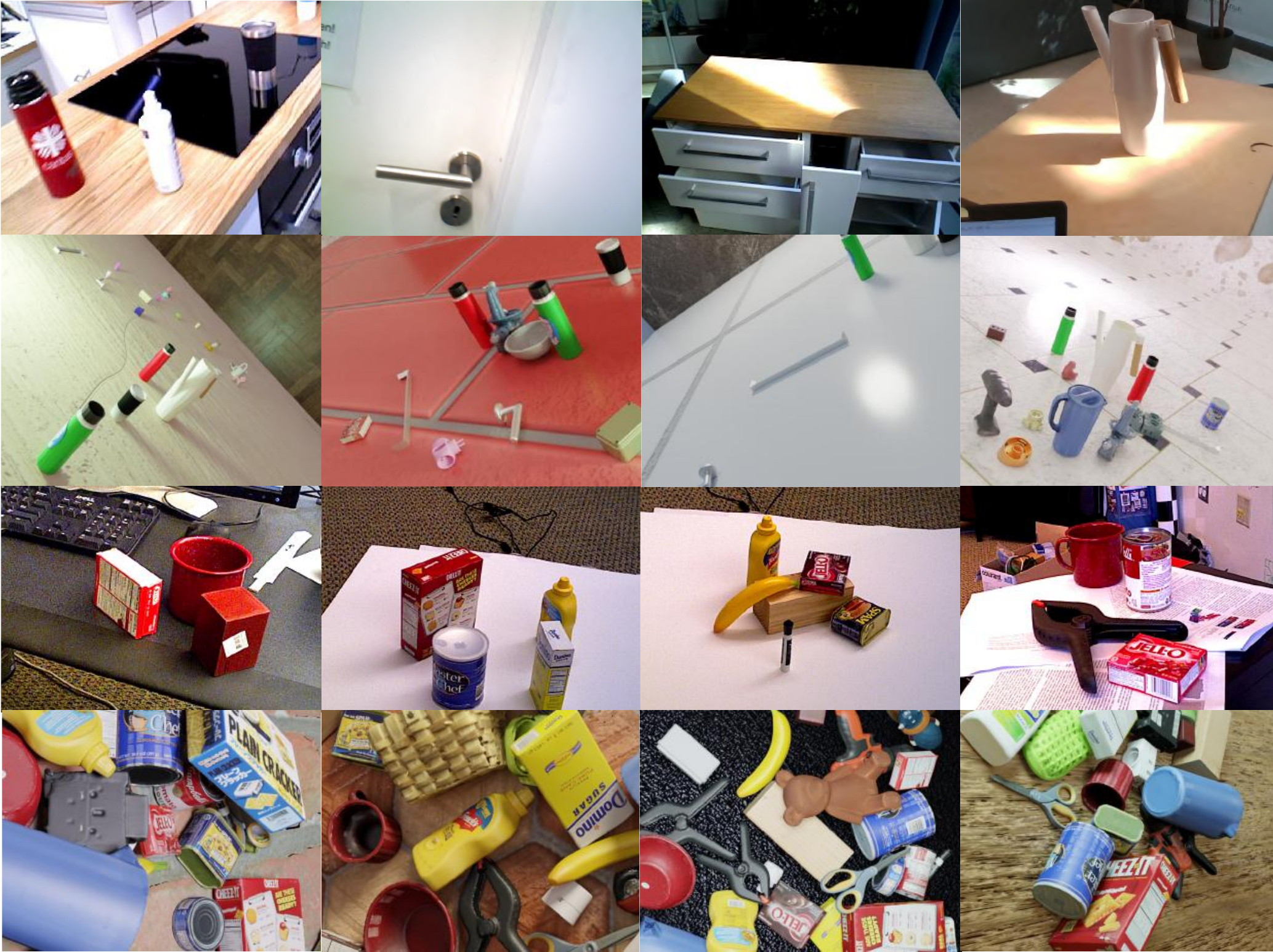}
	\end{center}
	\caption{\textbf{The real and synthetic data.} Exemplary images from real (1st,3rd row) and synthetic domain (2nd, 4th row) of EDAN (1-2 rows) and YCBV (3-4 rows) data sets.}
	\label{fig:images}
	\vspace*{-5mm}
\end{figure}

In this section, we first validate the proposed sampling strategy on a classification task, in which the model is transferred from MNIST \cite{lecun-mnisthandwrittendigit-2010} to MSNIST-M \cite{ganin2016domain}. 
Then we move on to two \textit{more challenging but task-relevant} self-collected data-sets on \textit{2D object detection}.
To note that, we employ two data-sets with different magnitudes of Sim-to-Real gap (one is large and the other small) to demonstrate that the proposed pipeline can efficiently bridge the gap for both cases.
In all experiments, \textit{we instantiate the Sim-to-Real gap by subtracting the performance of the corresponding models trained on purely the real and simulated data-set}.
Nevertheless, we address the limitation of the proposed idea by including one \textit{failure case} on the public YCBV data set~\cite{xiang2017posecnn} to further identify the operational scenario.
In the end, we show the practical effectiveness of our idea by deploying the model on an assistive robot within a grasping task.
The implementation details and parameter settings of the proposed pipeline are then provided, which is followed by results and discussions. 
\paragraph{Data sets}
(1) \textbf{Digits} include MNIST and MNIST-M digit data sets with \textit{10} classes. 
MNIST-M contains digits from MNIST but blended with random color patches.
We can treat MNIST-M as MNIST digits in real-world in this case and perform Sim-to-Real transfer for them. 
(2) \textbf{EDAN} includes \textit{5} classes: ikea bottle, watering can, door handle, drawer handle and grey mug.
With simple textures and geometry of the objects and the \textit{indoors lab environments} (see Fig. \ref{fig:images}), the domain gap on this data set is small.
(3) \textbf{SAM}~\cite{9197394} includes \textit{3} classes: cage, pipe and hook. 
With more complex textures and geometry of the objects and different weather conditions in \textit{outdoor environments}, the domain gap on this data set is much larger than EDAN.
(4) \textbf{YCBV} contains images of \textit{21} classes from common objects such as pitcher, sugar box and so on.
Basic information of the aforementioned data sets is summarized in Table \ref{table:data-set} and the synthetic data sets except for the one of 1 are generated by BlenderProc \cite{denninger2019blenderproc} with domain randomization applied.

\begin{table}[ht]
	\caption{Basic information and training hyper-parameters on four data sets}
	\vspace*{-8mm}
	\begin{center}
		\begin{tabular}{ |p{2.3cm}|p{0.9cm}|p{1.2cm}|p{1.2cm}|p{1.0cm}| }
			\hline
			Data set (size of sim, real-pool, real-val, real-test set, number of class)& Query Size (image) & Maximum Training Period during AL (epoch) & Learning Rate & Network Architecture\\
			\hline
			Digits data set ($60k$, $55k$, $5k$, $10k$, $10$) &  20 & 50  & linearly from $1e^{-5}$ to $1e^{-3}$& the same CNN in \cite{kirsch2019batchbald} \\
			\hline
			EDAN ($10k$, $0.5k$, $0.1k$, $1k$, $5$) &  20 & 10  & $1e^{-4}$ & RetinaNet \cite{lin2017focal} \\
			\hline
			SAM ($2.5k$, $2k$, $0.1k$, $0.5k$, $3$) &  80  &  10  & $1e^{-4}$ & RetinaNet \cite{lin2017focal} \\
			\hline
			YCBV ($50k$, $1.4k$, $0.1k$, $0.5k$, $21$) &  50  & 10 & $1e^{-3}$ & RetinaNet \cite{lin2017focal} \\
			\hline
		\end{tabular}
	\end{center}
	\label{table:data-set}
	\vspace*{-3mm}
\end{table}

\begin{table}[ht]
	\caption{Comparison of different acquisition functions: mean mAP over 10 iterations for different aggregation and combination functions with and without sub-sampling strategy and with 10 and 30 samples on EDAN data set.}
	\vspace*{-5mm}
	\begin{center}
		\begin{tabular}{ |p{0.8cm}|p{1.2cm}|p{1cm}|p{1cm}|p{1cm}|p{1cm}|  }
			\hline
			\multicolumn{2}{|c|} {} &\multicolumn{2}{|c|} {10 samples}&\multicolumn{2}{|c|} {30 samples}\\
			\hline
			Agg. & Comb. & w.o. sub. & w. sub. & w.o. sub. & w. sub.\\[1.5ex] 
			\hline\hline
			\multirow{2}{*}{Avg} & Max & 74.73\% & 76.51\% & 74.47\% & 76.76\%\\
			\cline{2-6}
			& Sum & 74.77\% & \textbf{77.09}\% & 75.17\% & \textit{77.19}\% \\
			\hline
			\multirow{2}{*}{Sum} & Max & 75.80\% & 74.54\% & 76.08\% & 76.76\%\\
			\cline{2-6}
			& Sum & 71.83\% & 75.35\% & 74.31\% & 76.67\% \\
			\hline
			\multirow{2}{*}{Max} & Max & 73.67\% & 72.67\% & 73.02\% & 74.74\%\\
			\cline{2-6}
			& Sum & 75.36\% & \textit{76.89}\% & 74.98\% & \textbf{77.49}\% \\
			\hline
		\end{tabular}
	\end{center}
	\label{table:design-choice}
	\vspace*{-5mm}
\end{table}

\paragraph{Baselines}
In order to validate the proposed idea, we compare with the following baselines.
(1) \textbf{Random}: an approach to randomly select data points for query in each iteration.
(2) \textbf{Batch-bald} \cite{kirsch2019batchbald}: an approach to query a batch of data with \textit{jointly maximum mutual information} instead of individually. 
(3) \textbf{Clue} \cite{prabhu2021active}: an approach for active domain adaptation that considers both \textit{diversity} and \textit{uncertainty} in the acquisition function.
(4) \textbf{Coreset} 
 \cite{sener2017active}: a \textit{diversity}-oriented approach for AL, whose greedy version is a k-center algorithm. 
For clue\footnote{https://github.com/virajprabhu/CLUE} and batch-batch\footnote{ https://github.com/BlackHC/BatchBALD}, we use the open-sourced implementation and only apply to 1 with $max$ aggregation function due to their iterative calculation characteristic.
For efficiency within coreset and clue, we use the logits layer as latent features.
\paragraph{Implementation details}

Training hyper-parameters are summarized in Table \ref{table:data-set}.
Within the $sum$ combination function, we set the weight of 1 to $1$ on all data sets. 
For regression, we select $0.01$ for EDAN and SAM, $0.001$ for YCBV.
The percentage of sub-sampling is set to 1\% for digits data sets and $50\%$ for the others based on the performance on validation set.
We set dropout rate to $0.1$ in BayesOD and apply Bayesian inference \textit{only for bounding box regression} instead of both heads to avoid under-performance observed in preliminary experiments. 
We use $100$ Monte-Carlo samples to approximate the joint distribution in batch-bald.
Regarding the evaluation metric, following the convention in \cite{prabhu2021active}, we employ the mean accuracies for classification and mean MAP for object detection over AL iterations. 

\subsection{Results and Analysis}
\paragraph{Design choices}
We firstly conduct an initial empirical study on the effects of aggregation and combination functions in Eq. \eqref{int_acq}, number of samples to approximate Eq. \eqref{eq:pred_dist} on EDAN data set (Table \ref{table:design-choice}).
We can observe: 1. More weight posterior samples can lead to slightly better results; 
\begin{figure}[ht]
	\includegraphics[width=\linewidth]{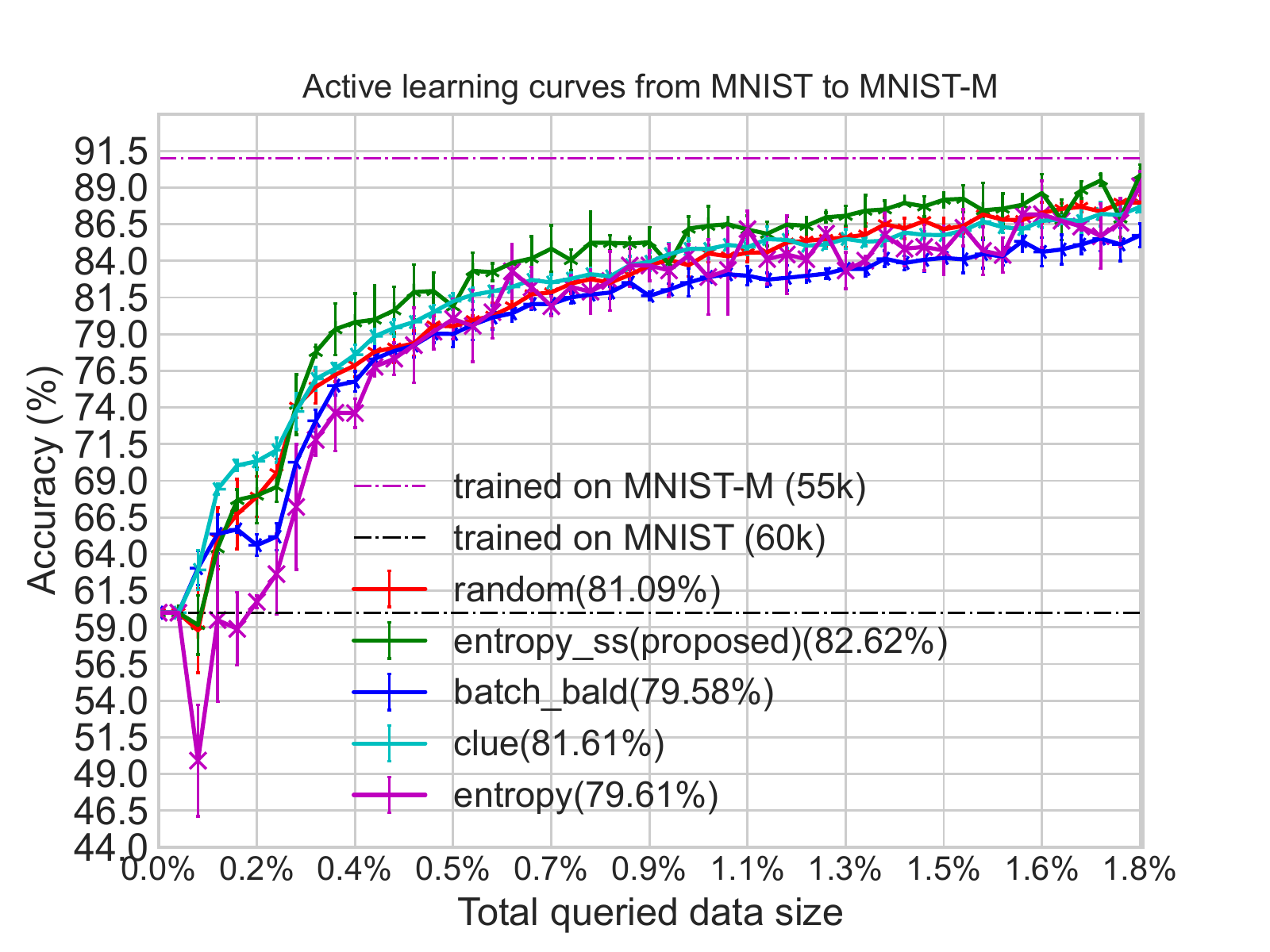}
	\caption{\textbf{Results on digits data sets (MNIST $\rightarrow$ MNIST-M).} Active learning learning curves of 3 random runs (with 50 iterations and 20 images queried in each iteration). The black and purple dotted lines represent the performance ( on MNIST-M test set) of model trained on MNIST and MNIST-M training set with size in the parentheses, respectively. The compared methods include the proposed one ($entropy\_ss$) and other baselines. Values in the parentheses are mean accuracies over 50 iterations. }
	\label{fig:digits_result}
	\vspace*{-1mm}
\end{figure}

\begin{table}[ht]	
	\caption{Results summary for object detection data sets. Values in this table are 1. \textbf{percentage of annotated images required to bridge Sim-to-Real gap} (lower the better) and 
		2. \textbf{mean mAP over 10 iterations within AL} (higher the better).}
	\vspace*{-5mm}
\begin{center}

	\begin{tabular}{ |p{1cm}|p{1cm}|p{1cm}|p{1cm}|p{1cm}|p{1cm}|p{1cm}| }
		\hline
		{} & Random & Proposed \ & Coreset & Clue & Batch-bald\\
		\hline
		EDAN &  $>40\%$ / \textit{75.7}\% & \textbf{36\%} / \textbf{77.1\%} & $>40$\% / 75.0\% & $>40\%$ / \textit{75.7}\% & $ >40\%$ /  72.9\%\\
		\hline
		SAM & $>40\%$ / 81.4\%  &  32\% / 82.2\% & \textbf{20\%} / \textbf{85.6\%} & 32\% / 85.0\% & $>40\%$ / 82.0\% \\
		\hline
		YCBV &  \textbf{40\%} / \textbf{65.2}\%  &   $>40\%$ / 63.5\% & \   $>40\%$ / 61.1\% & \textbf{40\%} / \textbf{65.2}\%&  $>40\%$ / 64.8\% \\
		\hline
	\end{tabular}
	\vspace*{-2mm}
\end{center}
\label{table:results}
\vspace*{-5mm}
\end{table}
2. The sub-sampling strategy can improve performance most of the cases; 
3. When using $avg$ and $max$ to aggregate uncertainties of detections on the image, the $sum$ combination function yields better results; Only within $sum$ aggregation function, the $max$ operation outperforms. 

In general, the setting pairs of $max+sum$ and $avg+sum$ provide the best results. 
As this ablates our design choices, we use this insight and mainly focus on these two settings with 10 samples and report only the one with better results.

\paragraph{Results on digits data sets}
In Fig. \ref{fig:digits_result}, we can see that the domain gap can be bridged with $\sim2\%$ data by the proposed sub-sampling strategy ($entropy\_ss$), faster and better than the $random$ and $clue$ baseline.
In contrast, the naive $entropy$ and $batch\_bald$ perform worse than $random$ along with large variations.
This shows that the proposed sampling strategy for mitigating distribution shift in AL is able to provide greater performance gain than the one considering trade-off between $uncertainty$ and $diveristy$.

\begin{figure}
	\includegraphics[ width=\linewidth]{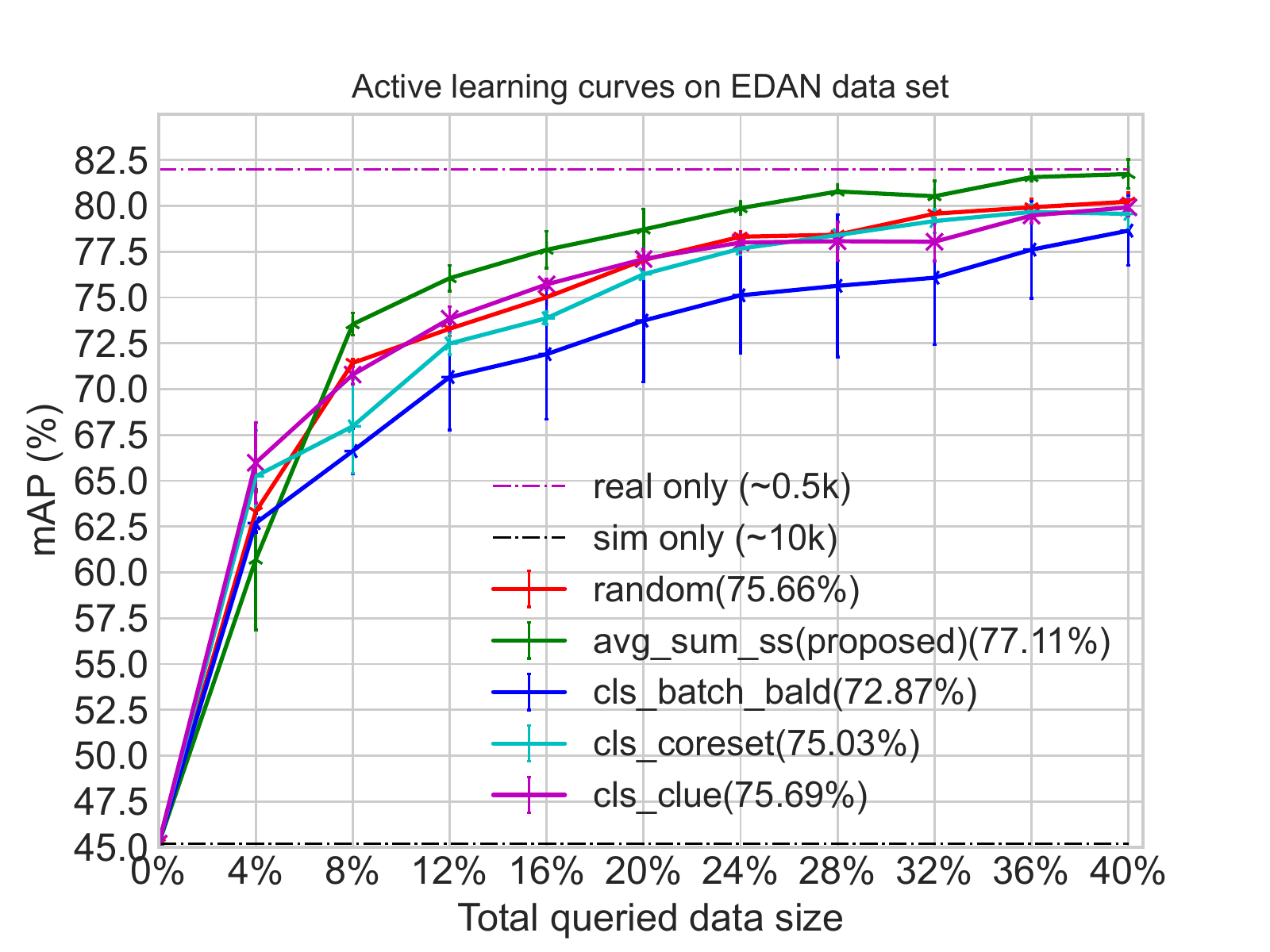}
	\caption{\textbf{Results on the EDAN data set.} Active learning curves of 3 random runs (with 10 iterations and 20 images queried in each iteration). The black and purple dotted lines represent the performance of model trained on sim and real data sets with size in the parentheses, respectively. The compared methods include the proposed one ($avg\_sum\_ss$) and other baselines. Values in the parentheses are mean mAP over 10 iterations.}
	\label{fig:edan_result1}
	\vspace*{-3mm}
\end{figure}

\paragraph{Results on EDAN data set}
In Fig. \ref{fig:edan_result1} and Table \ref{table:results}, we can learn that the gap can be eliminated by the proposed method with $avg$ to aggregate detections and $sum$ to combine $cls$ and $reg$ uncertainties($avg\_sum\_ss$) with only $36\%$ data, outperforming both the strong baseline $random$ and $clue$. 
In contrary, while $clue$ is on a pair with $random$ and slightly better than $coreset$, $batch\_bald$ has the lowest mean mAP.
This demonstrates that utilization of information from both $cls$ and $reg$ with sub-sampling is advantageous in the case of data set with moderate distribution shift like EDAN. 

\begin{figure}
	\includegraphics[ width=\linewidth]{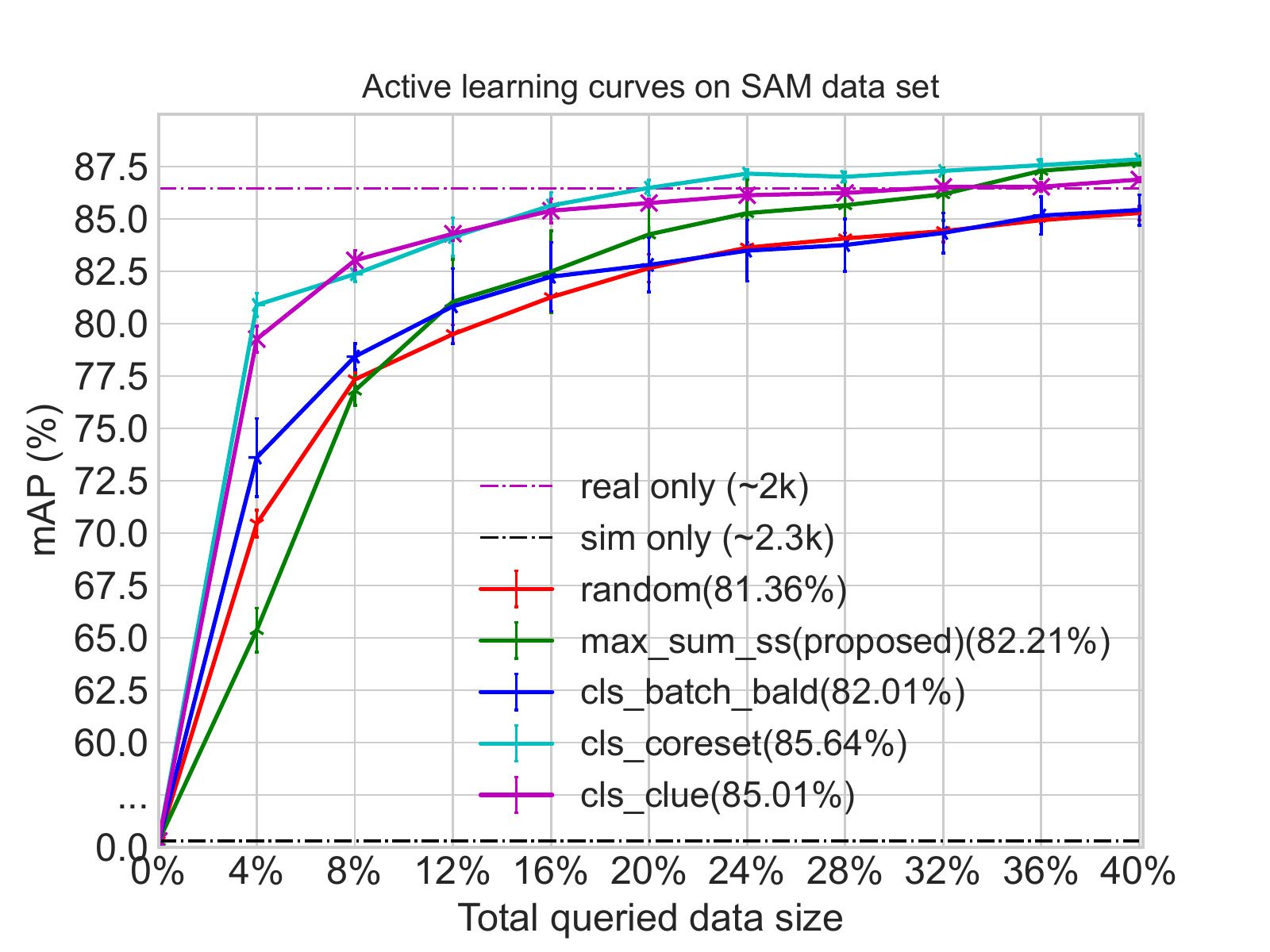}
	\caption{\textbf{Results on the SAM data set.} Active learning curves of 3 random runs (with 10 iterations and 80 images queried in each iteration). The black and purple dotted lines represent the performance of model trained on sim and real data sets with size in the parentheses, respectively. The compared methods include the proposed one ($max\_sum\_ss$) and other baselines. Values in the parentheses are mean mAP over 10 iterations.}
	\label{fig:sam_result1}
	\vspace*{-5mm}
\end{figure}

\paragraph{Results on SAM data set}
The final detector on this data set can achieve a quite decent mAP ($> 90\%$), therefore in this experiment we aim to bridge the gap up to a sufficient level, which is \textit{95\% of gap}.
In Fig. \ref{fig:sam_result1} and and Table \ref{table:results}, the proposed method with $max$ as aggregation and $sum$ as combination function followed by sub-sampling strategy ($max\_sum\_ss$) is still able to beat the strong $random$ baseline as well as $batch\_bald$ and diminish the gap.
Nevertheless, $clue$ and $coreset$ perform better than $max\_sum\_ss$ probably due to the larger domain gap on this data set.
It could attribute to the reason that our proposed sampling strategy aims to compensate the shift in label distribution, thereby less effective for a large shift in the input distribution.

\paragraph{Limitation Analysis}
In this sub-section, we show a failure case on YCBV data set to demonstrate the limitation of the proposed idea.
With this, we aim to identify the operational scenarios of AL for Sim-to-Real transfer and highlight the characteristic of this problem with the hope of providing some enlightening thoughts for the community. 

In the last row of Table \ref{table:results}, we see that all approaches are on a par with ($clue$) or worse than ($avg\_sum\_ss$, $coreset$, $batch\_bald$) the $random$ baseline.
To investigate the reason behind, inspired by \cite{aggarwal2020active}, we compute the average inter class variations over AL iterations in Table \ref{table:limitation}. 
The inter class variation is defined as the $\sigma \times C$, where $C$ is the number of class and $\sigma$ is the standard deviation of number of instances for all classes. 
The lower this value is, less variations and more balance the object category distribution possesses. 
We can quantitatively observe that variations of YCBV are significantly larger than the others due to greater number of class, which might pose greater difficulty on decreasing the label distribution shift.
Further from row-wise comparison, there is an obvious inversely proportional relation between inter class variations and the performance on YCBV, which is obscure on EDAN and SAM.  
Therefore, we infer that the impact of label distribution shift is more severe on data sets with greater number of class, thus impeding the effective utilization of uncertainty estimates. 
Considering this, we suggest that it is more effective to employ the proposed pipeline for bridging the reality gap when the class imbalance problem, one instance of label distribution shift is at a small scale.
\begin{table}[ht]
	\begin{center}
		\caption{Inter class variations for the selected data set in each iteration during active learning. Lower the better.}
		\begin{tabular}{ |p{1.0cm}|p{1.0cm}|p{1.0cm}|p{1.0cm}|p{1.0cm}|p{1.0cm}| }
			\hline
			{} & Random  & Sub-sampling \ & Core-set & Clue & Batch-bald\\
			\hline
			EDAN &  90  & 152 & \textbf{78} & 114 & 126\\
			\hline
			SAM & 81.3 &  39.9 & 71.8 & 75.3 & \textbf{12.6}\\
			\hline
			YCBV & \textbf{268} & 316 & 305 & \textbf{268} & 318\\
			\hline
		\end{tabular}
		\label{table:limitation}
	\end{center}
	\vspace*{-5mm}
\end{table}


\subsection{Deployment on EDAN}

\begin{figure}
	\includegraphics[ height=5.95cm, width=\linewidth]{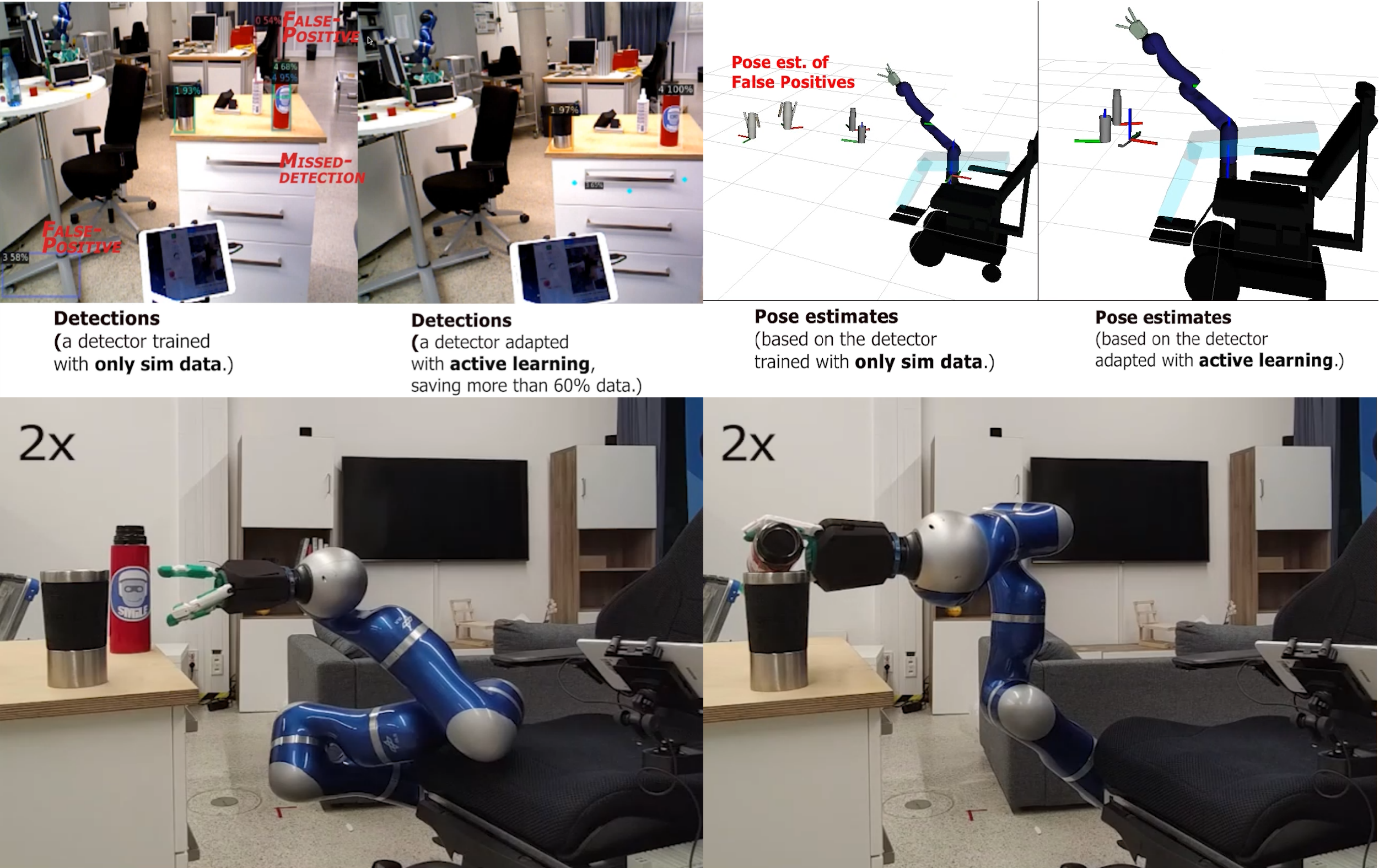}
	\caption{\textbf{Exemplary screenshots of a pouring task via shared control on an assistive robot.} The two screenshots on the top show the performance of the detector and the corresponding pose estimates (visualized in Rviz)  before (left in each column) and after (right in each column) adaptation via the proposed pipeline. The two screenshots at the bottom show the sequence of a grasping and pouring task execution with shared-control \cite{vogel2020edan}. }
	\label{fig:edan_demo}
	\vspace*{-3mm}
\end{figure}
On account of the working scenarios (e.g. care-giving) for an assistive robot \cite{vogel2020edan}, a variety of objects need to be detected and the manual efforts required for adaptation must be kept as minimum as possible.
Therefore, we show the effectiveness of the proposed idea in a shared-control grasping task on EDAN ( Fig. \ref{fig:edan_demo}), where a user such as people with motor disability sitting on the chair intends to control the robot arm for tasks like pouring by using an input device (EMG signal sensors or a spacemouse (used in the demo))with lower degrees of freedom (DoFs) than that of the end effector (3 vs. 6).
The mis-correspondence of DoFs between the input device and the manipulator demands that the user needs to tediously switch input mapping between them for task completion in a pure manual control mode.
In order to ease task execution, we employ shared-control templates \cite{quere2020shared}, which require robust and precise 2D object detection and pose estimation65. 
For more details on how to incorporate the perception pipeline into the shared-control module, we refer readers to the original work \cite{quere2020shared}.

In this demo, we integrate the adapted detector trained with a similar setting introduced in the previous section and further use the Agumented autoencoder (AAE) and Iterative Closest Point (ICP) pipeline \cite{sundermeyer2020augmented} for accurate pose estimation.
This perception pipeline can be deployed on an embedding system such as a NVIDIA Jetson TX2 or a workstation PC (used in demo), the predictions (i.e. pose estimates of the detected objects) are then sent to the shared-control module via Links and Nodes (LN) middle-ware. 
Based on this, the user is able to control the manipulator to perform a series of common daily tasks such as pouring and drinking with much less cognitive workload. 
We also provide a video to showcase the deployment.

%% file: chapters/conclusion.tex
\section{Conclusion}
This paper presents an active Sim-to-Real pipeline for 2D object detection, in which, a model is initially learned from synthetic data. 
Having observed the sub-optimal performance of learning only from simulation, we propose to efficiently use real annotated data via exploiting deep Bayesian active learning. 
Empirically, we demonstrate the encouraging impact of the proposed pipeline on classification and 2D object detection data sets, further address the limitation of the proposed pipeline and show its applicability on a real robotic system. 
In particular, our experiments indicate that the proposed sampling strategy can alleviate the label distribution shift which can have a vital impact on the success of our pipeline.
More importantly, our work provides an empirical evidence that the real annotated images can efficiently reduce the reality gap.

%% file: chapters/acknowledge.tex
\section{Acknowledgments}
We thank the anonymous reviewers and area chairs for their thoughtful feedback. 
Sincere thanks to the re-enabling robot (EDAN) team at DLR, especially the help on real robot deployment from Annette Hagengruber and Gabriel Quere. 
Jianxiang Feng is supported by the Munich School for Data Science (MUDS) and Rudolph Triebel is a member of MUDS.

%% file: root.bbl
\begin{thebibliography}{10}
\providecommand{\url}[1]{#1}
\csname url@rmstyle\endcsname
\providecommand{\newblock}{\relax}
\providecommand{\bibinfo}[2]{#2}
\providecommand\BIBentrySTDinterwordspacing{\spaceskip=0pt\relax}
\providecommand\BIBentryALTinterwordstretchfactor{4}
\providecommand\BIBentryALTinterwordspacing{\spaceskip=\fontdimen2\font plus
\BIBentryALTinterwordstretchfactor\fontdimen3\font minus
  \fontdimen4\font\relax}
\providecommand\BIBforeignlanguage[2]{{%
\expandafter\ifx\csname l@#1\endcsname\relax
\typeout{** WARNING: IEEEtran.bst: No hyphenation pattern has been}%
\typeout{** loaded for the language `#1'. Using the pattern for}%
\typeout{** the default language instead.}%
\else
\language=\csname l@#1\endcsname
\fi
#2}}

\bibitem{lin2017focal}
T.-Y. Lin, P.~Goyal, R.~Girshick, K.~He, and P.~Doll{\'a}r, ``Focal loss for
  dense object detection,'' in \emph{Int. Conf. on Computer Vision}, 2017.

\bibitem{durner_unknown_2021}
M.~Durner, W.~Boerdijk, M.~Sundermeyer, W.~Friedl, Z.-C. M{\'a}rton, and
  R.~Triebel, ``Unknown {Object} {Segmentation} from {Stereo} {Images},'' in
  \emph{{Int.} {Conf.} on {Intelligent} {Robots} and {Systems} ({IROS})}, 2021.

\bibitem{sundermeyer2020augmented}
M.~Sundermeyer, Z.-C. Marton, M.~Durner, and R.~Triebel, ``Augmented
  autoencoders: Implicit 3d orientation learning for 6d object detection,''
  \emph{Int. Journal of Computer Vision}, vol. 128, no.~3, 2020.

\bibitem{bousmalis2018using}
K.~Bousmalis, A.~Irpan, P.~Wohlhart, Y.~Bai, M.~Kelcey, M.~Kalakrishnan,
  L.~Downs, J.~Ibarz, P.~Pastor, K.~Konolige, \emph{et~al.}, ``Using simulation
  and domain adaptation to improve efficiency of deep robotic grasping,'' in
  \emph{ICRA}, 2018.

\bibitem{georgakis2017synthesizing}
G.~Georgakis, A.~Mousavian, A.~C. Berg, and J.~Kosecka, ``Synthesizing training
  data for object detection in indoor scenes,'' \emph{arXiv:1702.07836}, 2017.

\bibitem{10.1109/iros.2017.8202133}
J.~Tobin, R.~Fong, A.~Ray, J.~Schneider, W.~Zaremba, and P.~Abbeel, ``{Domain
  Randomization for Transferring Deep Neural Networks from Simulation to the
  Real World},'' \emph{IROS}, 2017.

\bibitem{denninger2019blenderproc}
M.~Denninger, M.~Sundermeyer, D.~Winkelbauer, Y.~Zidan, D.~Olefir,
  M.~Elbadrawy, A.~Lodhi, and H.~Katam, ``Blenderproc,''
  \emph{arXiv:1911.01911}, 2019.

\bibitem{muller_photorealistic_2021}
M.~G. M{\"u}ller, M.~Durner, A.~Gawel, W.~St{\"u}rzl, R.~Triebel, and
  R.~Siegwart, ``A {Photorealistic} {Terrain} {Simulation} {Pipeline} for
  {Unstructured} {Outdoor} {Environments},'' in \emph{IROS}, 2021.

\bibitem{Tanwani_DIRL_CORL_20}
A.~K. Tanwani, ``Domain invariant representation learning for sim-to-real
  transfer,'' in \emph{CoRL}, 2020.

\bibitem{electronics10121491}
M.~Ranaweera and Q.~H. Mahmoud, ``Virtual to real-world transfer learning: A
  systematic review,'' \emph{Electronics}, vol.~10, no.~12, 2021.

\bibitem{vogel2020edan}
J.~Vogel, A.~Hagengruber, M.~Iskandar, G.~Quere, U.~Leipscher, S.~Bustamante,
  A.~Dietrich, H.~H{\"o}ppner, D.~Leidner, and A.~Albu-Sch{\"a}ffer, ``Edan: An
  emg-controlled daily assistant to help people with physical disabilities,''
  in \emph{IROS}, 2020.

\bibitem{durner_experience-based_2017}
M.~Durner, S.~Kriegel, S.~Riedel, M.~Brucker, Z.~M{\'a}rton,
  F.~B{\'a}lint-Bencz{\'e}di, and R.~Triebel, ``Experience-based optimization
  of robotic perception,'' in \emph{{Int.} {Conf.} on {Advanced} {Robotics}
  ({ICAR})}, 2017.

\bibitem{feng2019introspective}
J.~Feng, M.~Durner, Z.-C. Marton, F.~Balint-Benczedi, and R.~Triebel,
  ``Introspective robot perception using smoothed predictions from bayesian
  neural networks,'' \emph{Robotics Research. ISRR 2019.}

\bibitem{gal2016dropout}
Y.~Gal and Z.~Ghahramani, ``Dropout as a bayesian approximation: Representing
  model uncertainty in deep learning,'' in \emph{Int. Conf. on Machine Learning
  (ICML)}, 2016.

\bibitem{harakeh2020bayesod}
A.~Harakeh, M.~Smart, and S.~L. Waslander, ``Bayesod: A bayesian approach for
  uncertainty estimation in deep object detectors,'' in \emph{Int. Conf. on
  Robotics and Automation (ICRA)}, 2020.

\bibitem{prabhu2019sampling}
A.~Prabhu, C.~Dognin, and M.~Singh, ``Sampling bias in deep active
  classification: An empirical study,'' \emph{arXiv:1909.09389}, 2019.

\bibitem{zhao2021active}
E.~Zhao, A.~Liu, A.~Anandkumar, and Y.~Yue, ``Active learning under label
  shift,'' in \emph{Int. Conf. on Artificial Intelligence and Statistics},
  2021.

\bibitem{su2020active}
J.-C. Su, Y.-H. Tsai, K.~Sohn, B.~Liu, S.~Maji, and M.~Chandraker, ``Active
  adversarial domain adaptation,'' in \emph{Winter Conf. on Applications of
  Computer Vision (WACV)}, 2020.

\bibitem{prabhu2021active}
V.~Prabhu, A.~Chandrasekaran, K.~Saenko, and J.~Hoffman, ``Active domain
  adaptation via clustering uncertainty-weighted embeddings,'' in \emph{CVPR},
  2021.

\bibitem{hinterstoisser2018pre}
S.~Hinterstoisser, V.~Lepetit, P.~Wohlhart, and K.~Konolige, ``On pre-trained
  image features and synthetic images for deep learning,'' in \emph{Europ.
  Conf. on Computer Vision (ECCV) Workshops}, 2018.

\bibitem{Photorealistic_Image_Synthesis}
T.~Hodan, V.~Vineet, R.~Gal, E.~Shalev, J.~Hanzelka, T.~Connell, P.~Urbina,
  S.~N. Sinha, and B.~Guenter, ``{Photorealistic Image Synthesis for Object
  Instance Detection},'' \emph{arXiv}, 2019.

\bibitem{zhu2019adapting}
X.~Zhu, J.~Pang, C.~Yang, J.~Shi, and D.~Lin, ``Adapting object detectors via
  selective cross-domain alignment,'' in \emph{Conf. on Computer Vision and
  Pattern Recognition (CVPR)}, 2019.

\bibitem{chen2018domain}
Y.~Chen, W.~Li, C.~Sakaridis, D.~Dai, and L.~Van~Gool, ``Domain adaptive faster
  r-cnn for object detection in the wild,'' in \emph{CVPR}, 2018.

\bibitem{cohn1996active}
D.~A. Cohn, Z.~Ghahramani, and M.~I. Jordan, ``Active learning with statistical
  models,'' \emph{Journal of Artificial Intelligence Research}, vol.~4, 1996.

\bibitem{feng2019deep}
D.~Feng, X.~Wei, L.~Rosenbaum, A.~Maki, and K.~Dietmayer, ``Deep active
  learning for efficient training of a lidar 3d object detector,'' in
  \emph{Intelligent Vehicles Symposium (IV)}, 2019.

\bibitem{kirsch2019batchbald}
A.~Kirsch, J.~Van~Amersfoort, and Y.~Gal, ``Batchbald: Efficient and diverse
  batch acquisition for deep bayesian active learning,'' \emph{Advances in
  Neural Information Processing Systems}, vol.~32, 2019.

\bibitem{wen2019bayesian}
J.~Wen, N.~Zheng, J.~Yuan, Z.~Gong, and C.~Chen, ``Bayesian uncertainty
  matching for unsupervised domain adaptation,'' \emph{arXiv:1906.09693}, 2019.

\bibitem{aghdam2019active}
H.~H. Aghdam, A.~Gonzalez-Garcia, J.~v.~d. Weijer, and A.~M. L{\'o}pez,
  ``Active learning for deep detection neural networks,'' in \emph{Int. Conf.
  on Computer Vision (ICCV)}, 2019, pp. 3672--3680.

\bibitem{roy2018deep}
S.~Roy, A.~Unmesh, and V.~P. Namboodiri, ``Deep active learning for object
  detection.'' in \emph{British Machine Vision Conf. (BMCV)}, vol. 362, 2018.

\bibitem{kao2018localization}
C.-C. Kao, T.-Y. Lee, P.~Sen, and M.-Y. Liu, ``Localization-aware active
  learning for object detection,'' in \emph{ACCV}, 2018.

\bibitem{choi2021active}
J.~Choi, I.~Elezi, H.-J. Lee, C.~Farabet, and J.~M. Alvarez, ``Active learning
  for deep object detection via probabilistic modeling,'' \emph{ICCV}, 2021.

\bibitem{lee2020estimating}
J.~Lee, M.~Humt, J.~Feng, and R.~Triebel, ``Estimating model uncertainty of
  neural networks in sparse information form,'' in \emph{ICML}, 2020.

\bibitem{miller2018dropout}
D.~Miller, L.~Nicholson, F.~Dayoub, and N.~S{\"u}nderhauf, ``Dropout sampling
  for robust object detection in open-set conditions,'' in \emph{ICRA}, 2018.

\bibitem{lee2022}
J.~Lee, J.~Feng, M.~Humt, M.~G. M\"uller, and R.~Triebel, ``Trust your robots!
  predictive uncertainty estimation of neural networks with sparse gaussian
  processes,'' in \emph{5th Annual Conference on Robot Learning (CoRL)}, 2021.

\bibitem{gawlikowski2021survey}
J.~Gawlikowski, C.~R.~N. Tassi, M.~Ali, J.~Lee, M.~Humt, J.~Feng, A.~Kruspe,
  R.~Triebel, P.~Jung, R.~Roscher, \emph{et~al.}, ``A survey of uncertainty in
  deep neural networks,'' \emph{arXiv:2107.03342}, 2021.

\bibitem{hall2020probabilistic}
D.~Hall, F.~Dayoub, J.~Skinner, H.~Zhang, D.~Miller, P.~Corke, G.~Carneiro,
  A.~Angelova, and N.~S{\"u}nderhauf, ``Probabilistic object detection:
  Definition and evaluation,'' in \emph{WACV}, 2020.

\bibitem{mackay1992information}
D.~J. MacKay, ``Information-based objective functions for active data
  selection,'' \emph{Neural Computation}, vol.~4, no.~4, 1992.

\bibitem{yang2010ensemble}
Y.~Yang, G.~Ma, \emph{et~al.}, ``Ensemble-based active learning for class
  imbalance problem,'' \emph{Journal of Biomedical Science and Engineering},
  vol.~3, no.~10, 2010.

\bibitem{aggarwal2020active}
U.~Aggarwal, A.~Popescu, and C.~Hudelot, ``Active learning for imbalanced
  datasets,'' in \emph{WACV}, 2020.

\bibitem{oksuz2020imbalance}
K.~Oksuz, B.~C. Cam, S.~Kalkan, and E.~Akbas, ``Imbalance problems in object
  detection: A review,'' \emph{Trans. on Pattern Analysis and Machine
  Intelligence}, 2020.

\bibitem{lecun-mnisthandwrittendigit-2010}
Y.~LeCun and C.~Cortes, ``{MNIST} handwritten digit database,'' 2010.

\bibitem{ganin2016domain}
Y.~Ganin, E.~Ustinova, H.~Ajakan, P.~Germain, H.~Larochelle, F.~Laviolette,
  M.~Marchand, and V.~Lempitsky, ``Domain-adversarial training of neural
  networks,'' \emph{Journal of Machine Learning Research}, vol.~17, no.~1,
  2016.

\bibitem{xiang2017posecnn}
Y.~Xiang, T.~Schmidt, V.~Narayanan, and D.~Fox, ``Posecnn: A convolutional
  neural network for 6d object pose estimation in cluttered scenes,''
  \emph{Robotics: Science and Systems (RSS)}, 2018.

\bibitem{9197394}
J.~Lee, R.~Balachandran, Y.~S. Sarkisov, M.~De~Stefano, A.~Coelho, K.~Shinde,
  M.~J. Kim, R.~Triebel, and K.~Kondak, ``Visual-inertial telepresence for
  aerial manipulation,'' in \emph{ICRA}, 2018.

\bibitem{sener2017active}
O.~Sener and S.~Savarese, ``Active learning for convolutional neural networks:
  A core-set approach,'' \emph{Int. Conf. on Learning Representations (ICLR)},
  2018.

\bibitem{quere2020shared}
G.~Quere, A.~Hagengruber, M.~S.~Z. Iskandar, S.~Bustamante~Gomez, D.~Leidner,
  F.~Stulp, and J.~Vogel, ``Shared control templates for assistive robotics,''
  in \emph{ICRA}, 2020.

\end{thebibliography}
